# Defining and Explorting[1] the Intelligence Space

Paul S. Rosenbloom
Thomas Lord Department of Computer Science
University of Southern California
*Draft of June 16, 2023*

**Abstract**

Intelligence is a difficult concept to define, despite many attempts at doing so. Rather than trying to settle on a single definition, this article introduces a broad perspective on what intelligence is, by laying out a cascade of definitions that induces both a nested hierarchy of three levels of intelligence and a wider-ranging space that is built around them and approximations to them. Within this intelligence space, regions are identified that correspond to both natural – most particularly, human – intelligence and artificial intelligence (AI), along with the crossover notion of humanlike intelligence. These definitions are then exploited in early explorations of four more advanced, and likely more controversial, topics: the singularity, generative AI, ethics, and intellectual property.

Consider the space of all possible forms of intelligence, what can be called the *intelligence space*. We know very little about the nature of this space other than how it is currently populated with various, often only partially understood, instances of natural and artificial intelligence, although some earlier thoughts on this space – typically under different names – can be found in [2-5]. Rosenbloom [6] takes a different tack at understanding this space, in attempting to define a set of dichotomies whose cross product spans technologies underlying artificial and human intelligence, with the possibility of it spanning a much larger swath of the intelligence space.

Here, another tack is taken, of starting with a generic, trilevel definition of intelligence that anchors the space and hypothesizing that a range of approximations to this definition could flesh out much of the full space (Section 1). This space is then exploited by exploring the relationship between natural and artificial intelligence via the mapping of human intelligence, humanlike intelligence, artificial intelligence, and cognitive science onto regions of this space (Section 2) and then beginning an exploration into the implications of this all for four more advanced, and likely more controversial topics: the singularity, as it relates to intelligence; ethics, with a particular focus on its relationship to artificial intelligence (AI); the current white-hot topic of generative AI, with a particular focus on large language models (LLMs); and intellectual property, with a particular focus on whether or not its creation might be ascribed to AI systems (Section 3). Section 4 then concludes.

Structurally, this article proceeds via a cascade of concepts and their definitions, in the spirit of Lucretius [7] and Spinoza [8], along with associated commentary when appropriate.[2] Although the scope of this effort is not nearly as wide ranging as either of these previous ones, it does share with them the attempt to reconstruct, and to re-understand, an important range of complex concepts from the ground up. Like them as well, it includes topics that may be controversial, particularly the more advanced ones in Section 3.

---

[1] According to the *Urban Dictionary* (2016), explortation combines exploration and exploitation. The term originally had negative connotations in the context of colonialism, but it does yield a useful portmanteau for a form of scientific or philosophical investigation.

[2] A similarity in spirit can also be discerned with respect to a recent blog post by Yoshua Bengio [9] that provides a sequence of "definitions, hypotheses, and resulting claims about AI systems."

The overall approach bears a resemblance to how mathematics builds up from simple to complex concepts – Spinoza, in fact, appeals to the method of Euclid [10] – but it is less formal, with no attempt at articulating theorems and proofs. It also bears a resemblance to modern work on ontologies, and in this spirit is illustrated with numerous dependency diagrams among related sets of concepts; yet, in reality it is more attuned to the earlier philosophical efforts already mentioned. So, this may best be considered as part of a broad philosophy of mind that is aimed at a wide range of possible minds, both natural and artificial. It is written, however, from the perspective of someone who, rather than being a professional philosopher, has been active in artificial intelligence and cognitive science over many decades and who has more recently taken to reflecting extensively on this experience.

This work occurs in a certain sense in isolation, driven by its own internal consistency and momentum, rather than being a deeply scholarly work concerned with relating the definitions that result to everything relevant out there, although such contact is made when particularly helpful. It is not the originality of the definitions – to the extent that they are original – nor even that the definitions are exactly right or that the terms associated with them are necessarily the right terms, that matters, but that a step forward is made in understanding intelligence and its overall space of possibilities that provides a coherent skeleton for jointly framing natural and artificial intelligence while providing insight into some of the more speculative and controversial, yet crucial, issues concerning them.

# 1 Intelligence and the Intelligence Space

There is no single, consensus definition of intelligence. Legg [11] lists 71 different definitions of intelligence, while quoting Robert J. Sternberg as saying that "Viewed narrowly, there seem to be almost as many definitions of intelligence as there were experts asked to define it." Here, the question of how to define intelligence is approached via a nested hierarchy of three levels of intelligence. The most minimal form is *immediate intelligence*, which concerns acting rationally in the here and now (Section 1.1). *Cumulative intelligence* then adds to this the ability to improve over a lifetime via learning (Section 1.2). *Full-spectrum intelligence* takes a more comprehensive perspective by adding in much more of the breadth implicated in human intelligence, including situatedness, (self-)awareness, projection of future possibilities, generality across problems, feelings, purposefulness, language, collaboration, socialization, and education (Section 1.3).

Intelligence, broadly construed, can then concern any of these three levels – each essentially specifying an ideal – as well as approximations of them (Section 1.4). The end result is thus not a single compact definition but a rich space of possible forms of intelligence – or, in other words, what can be viewed as an *intelligence space* – that may be considered when exploring forms of both natural and artificial intelligence. A key result from this analysis is that existing definitions of intelligence are not all concerned with the same thing, but that this hierarchy can help to make sense of this diversity.

## 1.1 Immediate Intelligence

Immediate intelligence concerns the making of rational decisions in the here and now; in particular, how to choose actions that achieve goals. This section proceeds to define immediate intelligence across four steps. The first three subsections introduce a hierarchy of entity types, starting with *objects*, continuing through *systems* and *beings*, and ultimately ending with *agents*. The fourth subsection then defines immediate intelligence in terms of the mind of a rational agent. This section ultimately aligns reasonably well with the corresponding parts of Chapter 2 of Russell & Norvig [12], although not all of the terminology is the same.

*1.1.1 System (Figure 1)*
Here we begin with the minimal concept of an object and build up from there to that of a system that is based on mechanism(s). The notion of an architecture, as a system's fixed mechanisms is also highlighted.

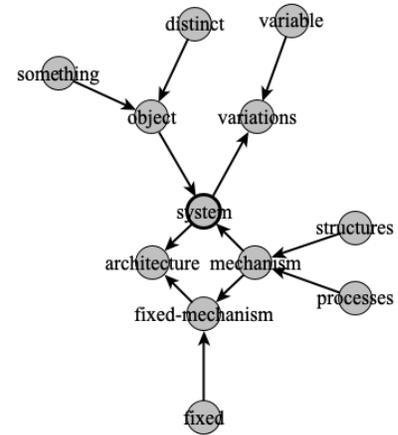

1. An *object* is a distinct something.
   - It need not be physical.
   - This aligns with one key aspect of a computational symbol described in Rosenbloom [13]. However, the other aspect – indecomposability – need not apply to objects here.

Figure 1: Dependency structure for a System.

2. A *mechanism* is a combination of structures and processes.
   - This aligns with the notion of science, writ large, as concerned with understanding the interactions among particular structures and processes [14].
3. A *fixed mechanism* remains unchanged over time.

4. A *system* is an object that includes one or more mechanisms.
   - According to Rosenbloom [15], a system can be decomposed semi-formally as system = architecture + variations.
5. The *architecture* comprises the fixed mechanisms of the system.
6. The *variations* comprise the aspects of the system that may change over time.

*1.1.2 Being (Figure 2)*
Given the notion of a system, we build up here to the concept of a being, which can be either natural or artificial. Introduced in the process is one sense of how a being's internal mechanisms can be grounded in what they represent [16].

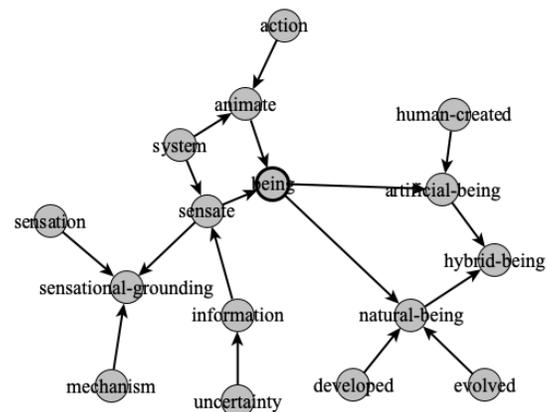

7. An *animate* is a system capable of external action.

8. *Information* reduces uncertainty.
9. A *sensation* is information that becomes available to a system.

Figure 2: Dependency structure for a Being.

   - This largely concerns external information but may also be internal.
10. A *sensate* is a system capable of processing sensations.

11. *Sensational grounding* ties the internal mechanisms of a sensate to the sensations that engender them.
    - This is one approach to the problem of symbol grounding. A second approach can be found in Section 1.3.2.
    - This can also be considered as providing a procedural form of grounding, as it centers on the processing that relates sensations to internal mechanisms.

12. A *being* is a system that is both animate and sensate.
    - Assumes some form of embodiment in being both animate and sensate.
    - The study of beings draws on disciplines such as biology and engineering.

13. A *natural being* has evolved and develops.
    - Development may slowly change a being's architecture.
14. An *artificial being* has been created (not birthed) by a natural being.
    - This distinction between natural and artificial can in principle also apply to objects, systems, animates, and sensates; as well as to the agents introduced shortly.
    - Section 3.1 extends this to artificial beings created by other artificial beings.
15. A *hybrid being* mixes properties of natural and artificial beings.

## 1.1.3 Agent (Figure 3)

Here the focus shifts up to agents – beings with minds that include goals to be achieved and memories. Brains are introduced as systems that control beings and implement their minds, and cognitive architectures as the fixed structures of these minds.

16. A *brain* is a system that controls a being.

17. A *mind* is a functional system implemented by a brain.
    - In being functional, minds concern behavioral results at some level of abstraction.
18. A *cognitive being* is a being with a mind.
19. A *cognitive architecture* is the architecture of a mind.

20. A *goal* is something desired.
    - This is intended to be a broad notion that includes achieving particular states, abiding by particular norms or constraints, meeting particular standards, maximizing particular rewards or objective functions, accomplishing particular missions, etc. It also includes any background motivations that lead to particular goals.

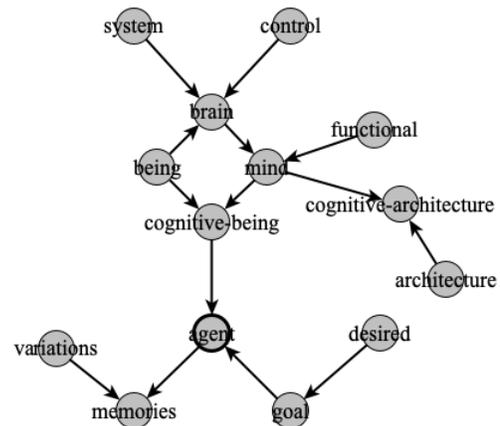

Figure 3: Dependency structure for an Agent.

21. An *agent* is a goal-oriented – i.e., teleological – cognitive being.
    - Goals may be explicit or implicit within the agent.

- All natural beings are assumed to be agents, although many may only have implicit goals.

22. *Memories* are the variations associated with an agent.

*1.1.4 Immediate Intelligence (Figure 4)*
An *essential definition* of a concept strips out from existing definitions all but its most essential aspects, leaving what remains as no longer definitional yet still useful in helping to define a space of meaningful variations over the concept [17]. Immediate intelligence, based on the mind of a rational agent, suggests such an essential definition for intelligence. Key elements of the space of variations this definition enables can be found in Sections 1.2 and 1.3, in terms of cumulative intelligence and full-spectrum intelligence.

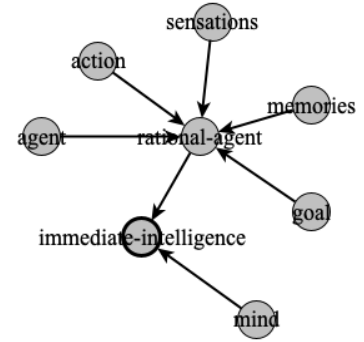

Figure 4: Dependency Structure for Immediate Intelligence

23. A *rational agent* chooses the best actions given its sensations, memories, and goals.
    - The study of rational intelligence draws on disciplines such as artificial intelligence – most particularly, knowledge representation and reasoning – decision theory, and cognitive psychology.
24. *Immediate intelligence* is what is yielded by the mind of a rational agent.
    - May be realized at various levels of approximation across different agents.

## 1.2 Cumulative Intelligence (Figure 5)

Cumulative intelligence adds lifelong learning to what is already provided by immediate intelligence. This goes significantly beyond the short-term rationality that underlies immediate intelligence but could potentially be considered as a long-term form of rationality that is concerned not just with the here and now but also with future decisions.

Machine learning – that is, learning in artificial systems – has been the dominant topic in AI for a number of years, and in the process has made a strong case for the criticality of learning in achieving intelligence, further highlighting the need to distinguish cumulative intelligence from both immediate and full-spectrum intelligence, even though conventional machine learning yields only an approximation to lifelong learning due to its focus on pretraining rather than online learning over a lifetime.

25. *Learning* turns sensations into memories.
    - May relate sensations to each other and to existing memories in the process.
    - May acquire or modify non-architectural mechanisms and and/or their parameters.
26. *Machine learning* is learning by an artificial being.
    - A sensate can in principle learn but the focus is on the more general case of a being here.

27. *Online learning* occurs incrementally, in parallel with the stream of sensations being received.

28. *Lifelong learning* is online learning across the lifetime of a being.

29. *Pretraining* completes all learning before "real" performance begins.
    - This is how most modern machine learning proceeds – except for reinforcement learning [18] – but it provides only an approximation to true lifelong learning.

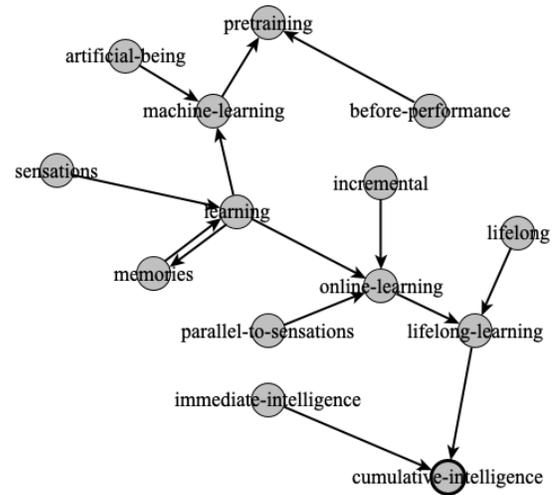

Figure 5: Dependency structure for Cumulative Intelligence.

30. *Cumulative intelligence* comprises immediate intelligence plus lifelong learning.
    - May be realized at various levels of approximation across different beings.
    - Draws on disciplines such as (machine) learning, reinforcement learning, optimization, cognitive architecture, and (cognitive) neuroscience.
    - "Reward is enough" [19] implies reinforcement learning is sufficient for intelligence. Such approaches as Universal AI [20] and deep reinforcement learning [21] say that an additional form of machine learning, in the form of induction, is also needed.

**1.3 Full-spectrum Intelligence**

Full-spectrum intelligence starts with cumulative intelligence but then adds to it a list of other aspects/capabilities presumed to be relevant for achieving the kind of intelligence found in humans (see Section 2.1 for more on human intelligence). Such a list is inherently ad hoc to some degree, but the version here is a "best attempt" based on contemplating previously published lists from researchers in cognitive architecture [22-26] and artificial general intelligence (AGI) [27]. Cognitive architecture studies the architecture of mind(s), whereas artificial general intelligence studies how to (learn to) perform (at least) any task performable by humans. Both areas are thus deeply concerned with the full spectrum of what is necessary for intelligent behavior.

Also included in the deliberations has been recent evidence concerning the capabilities and limitations of state-of-the-art neural networks – particularly those that learn to play games (e.g., [28]) and large language models (e.g., [29]) – plus a definition of intelligence I developed some years ago while teaching introductory courses on AI: "The common underlying capabilities that enable a system to be general, literate, rational, autonomous and collaborative." The overall result of these deliberations focuses on agents that are not only cumulative, but also situated, aware, projective, affective, purposeful, linguistic, collaborative, socialized, and educated.

*1.3.1 Situated Agent (Figure 6)*
There has been controversy over the years in AI and cognitive science as to whether intelligence can be studied in the abstract or whether it is inherently tied to the context in which an agent is situated (e.g., [30]). Either way, when an agent is actually situated in an environment, the latter does place constraints on the former. The definition of a situated agent here does not include all aspects mentioned in the literature on situatedness, such as culture, but it is intended to cover the most essential aspect, with culture being considered separately, as part of the discussion of collaborative agents in Section 1.3.8.

31. A *situated agent* can behave appropriately with respect to its goals in an environment that imposes constraints on sensation and animation.

32. A *bot* is an artificial situated agent.
33. A *virtual human* is a virtual bot with a humanlike appearance.
34. A *robot* is a physical bot.
35. An *android* is a robot with a humanlike appearance.

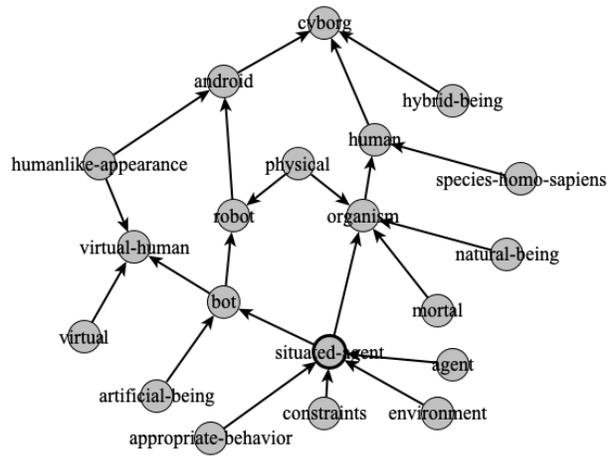

Figure 6: Dependency structure for a Situated Agent.

36. An *organism* is a natural, mortal, physically situated agent.
    - Although organisms can also participate in virtual environments through appropriate interfaces.
37. A *human* is an organism that is a member of the species Homo sapiens.

38. A *cyborg* is a hybrid being that is part human and part android.

*1.3.2 Awareness (Figure 7)*
The definitions here attempt to walk a fine line of introducing important aspects of tricky topics such as understanding, awareness, and self-awareness while avoiding even trickier topics such as consciousness.

39. A *model* is memories that are about something.
40. *Meaning* is the relationship between a model and what it models.
41. *Meaningful grounding* ties models to what they are about.

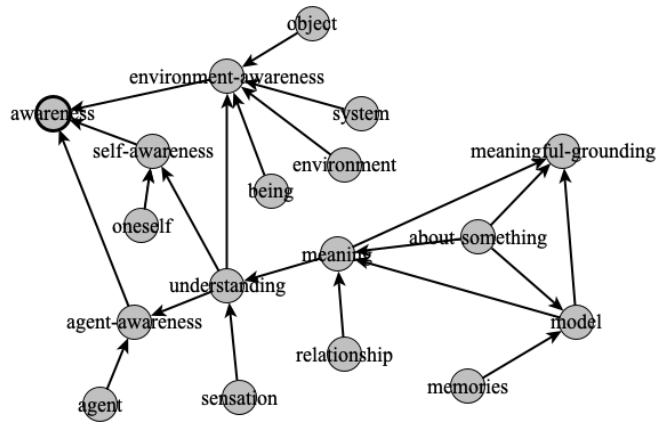

Figure 7: Dependency structure for Awareness.

   - This goes beyond sensational grounding but almost necessarily arises from it.
   - It can also be considered as providing a declarative form of grounding that complements the procedural form yielded by sensational grounding.
42. *Understanding* is a sensation of meaning.
   - Requires sensations of models, of what is modeled, and of the relationship between them.

43. *Self-awareness* is understanding of oneself.
    - The ability to sense oneself, have a model of oneself that can be sensed, and sense the relationship between oneself and the model of oneself.
    - This need not imply full self-awareness in which one can understand everything about oneself but does imply some minimal level of it.
    - This relates to concepts of reflection (e.g., [31]) and metacognition (e.g., [32]) but these two terms are not otherwise used here.
44. *Environment-awareness* is understanding of objects, systems, and beings in the environment.
45. *Agent-awareness* is understanding of other agents.
    - This includes Theory of Mind for understanding the minds of other agents.
46. *Awareness* is a combination of self-awareness, environment-awareness, and other-awareness.

### 1.3.3 Projective Agent (Figure 8)

Projection is a basic capability that underlies anticipating the future, as found for example in search, problem solving, and planning. It can be crucial in appropriately responding to unexpected situations. It is also crucial to creativity in hypothesizing what does not already exist.

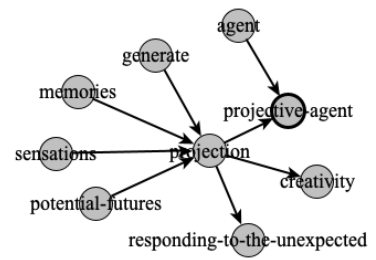

Figure 8: Dependency structure for a Projective Agent.

47. *Projection* uses memories and sensations to generate and sense potential futures.
48. A *projective agent* is capable of projection.

### 1.3.4 General Agent (Figure 9)

Generality is the heart of artificial general intelligence and is a key driver of work on cognitive architectures.

49. A *general agent* can work appropriately on a broad range of goals.
    - This is typically conceived of as being comparable to the set of goals on which a human can work.

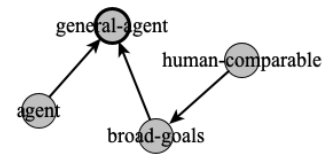

Figure 9: Dependency structure for a General Agent.

### 1.3.5 Affective Sensate (Figure 10)

The notion here of "truly matters" is intended to capture the essence of visceral feelings, such as pain and pleasure, in natural beings that has evolved to induce a particular form of urgency that is difficult, if not nearly impossible, to ignore. I previously tried to capture aspects of this in work on emotions in the Sigma cognitive architecture by ensuring that there was an architectural aspect to them that goes beyond simple reasoning about emotions – by, for example, changing how the system thinks [33] – but this still falls short of feelings. The difference likely has to do with aspects of physiology in natural beings, but a general definition of what this is, and how it might thus relate to artificial beings, remains unclear.[3] Whatever the answer is, feelings of this sort seem particularly relevant to constructing an ethical framework that is based on aspects of beings that are more elemental than their species membership (Section 3.3).

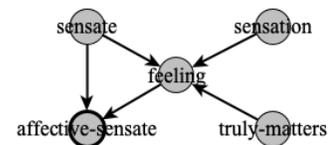

Figure 10: Dependency structure for an Affective Sensate.

50. A *feeling* is a sensation that "truly matters" to a sensate.
    - Distinguishes representing and thinking about emotions from experiencing them.

51. An *affective sensate* is a sensate that has feelings.
    - At its simplest, this might just be pleasure (positive feelings) and pain (negative feelings).

*1.3.6 Purposeful Agent (Figure 11)*
Purposes go beyond goals to aims that "truly matter." They are akin to feelings in this way and are similarly an essential component of a deeper ethical framework (Section 3.3).

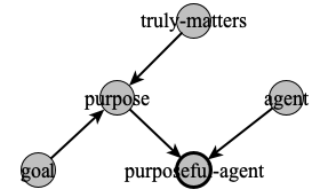

Figure 11: Dependency structure for a Purposeful Agent.

52. A *purpose* is a goal that "truly matters" to an agent.
53. A *purposeful agent* is an agent that has purposes.

*1.3.7 Linguistic Being (Figure 12)*
A linguistic being uses language to communicate with others.

54. A *listening being* can receive information generated by an animate.
55. A *talking being* can generate information for a sensate.
56. A *communicative being* is both a listening being and a talking being.
57. *Language* is communication that is understood.
58. A *linguistic being* is a communicative-being that uses language.

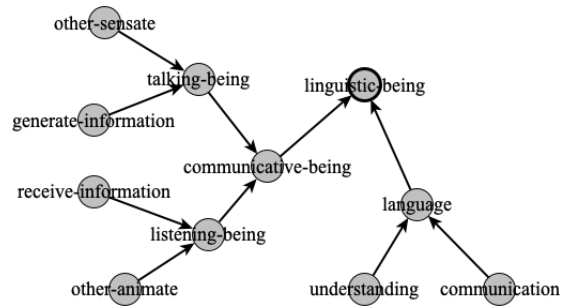

Figure 12: Dependency structure for a Linguistic Being.

*1.3.8 Collaborative Agent (Figure 13)*
A collaborative agent shares goals with others in an organization. The definitions here start with the notion of an aggregation of beings, and of a culture as its architecture, and works up from there.

59. An *aggregation* is a collection of beings.
60. A *culture* is the architecture of an aggregation.
    - It evolves over the history of the aggregation.

61. An *organization* is an aggregation of agents that is itself an agent.

---

[3] This is not intended as an appeal to *vitalism*, the notion that there is some non-physical vital force that distinguishes living beings from everything else [34], as the assumption here is that there is a physical, likely chemical, explanation for the phenomenon.

62. A *human organization* is an organization of humans and is typically a hybrid agent.
    - While humans are natural, the organization itself is created by humans and thus is not completely natural.

63. A *shared goal* is a goal of an organization.
64. A *collaborative agent* is an agent that has a shared goal.

*1.3.9 Socialized Agent*

A socialized agent is aware of its situation within a society and behaves appropriately within such a context. In a sense, a socialized agent can be considered simply as an aware, situated agent within a social environment. We will build up to the notion of a society through the introduction of a number of intermediary concepts, such as rights, responsibilities, autonomy, and authority. Due to the overall complexity of the dependency graph for a socialized agent (Figure 15), it is presented in two parts, with the first focused just on the notion of a right (Figure 14). Further discussion of rights and responsibilities can be found in Section 3.3.8, as part of a broader discussion of ethics (Section 3.3).

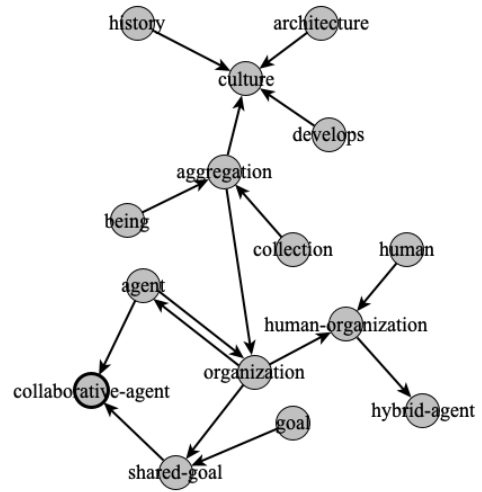

Figure 13: Dependency structure for a Collaborative Agent.

65. A *benefit* provides outside assistance to an object.
    - Benefits may, for example, aid the existence of an object, a feeling of an affective sensate, or a goal of an agent.

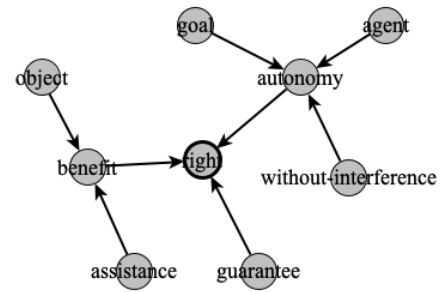

Figure 14: Dependency structure for a Right.

66. *Autonomy* is the ability of an agent to choose its own goals and to pursue them without outside interference.

67. A *right* is a guarantee.
    - It might guarantee a level of autonomy with respect to an agent's goals or a level of benefit to an object.
    - Individual rights may require tradeoffs when they conflict with other rights.

68. A *responsibility* is an obligation by an agent to a goal.

69. An *authority* is an agent that can determine rights/responsibilities for objects/agents.

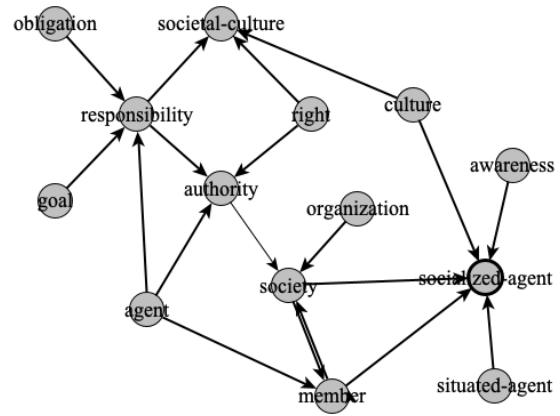

Figure 15: Dependency structure for a Socialized Agent.

70. A *society* is an organization with authority over its members.
71. A *member* is an agent within a society over which the society has authority.
72. A *societal culture* includes the rights and responsibilities associated with membership.
73. A *socialized agent* is a situated member of a society that has awareness of the society's culture and acts appropriately within this culture.

*1.3.10 Educated Agent (Figure 16)*
In contrast to learning, which focuses on how sensations are turned into memories, education focuses on how what society as a whole has learned – and curated in the sense of having reached a consensus on its validity – is reflected in an agent. It may be learned by the agent in some manner, or it could potentially be embedded into an agent via other means, such as evolution, development, or programming. Either way, the result should be curated memories. There is a conventional – and essentially correct – notion that ignorance is distinct from lack of intelligence, but the lack of education on the part of generative AI systems highlights how much education matters to overall judgments of an agent's intelligence. From the perspective here, ignorance is distinct from a lack of immediate intelligence but not from a lack of full-spectrum intelligence. Cumulative intelligence exists in a murkier intermediate ground, in which an agent can be considered lacking if its memories don't reflect its experiences.

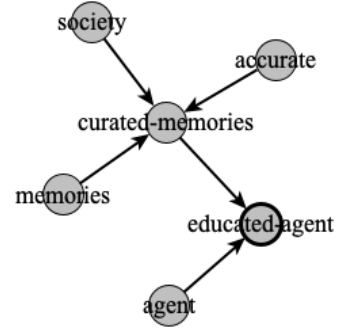

Figure 16: Dependency structure for an Educated Agent.

74. *Curated memories* have been determined by a society to be accurate.
75. An *educated agent* has curated memories.

*1.3.11 Full-spectrum Intelligence (Figure 17)*
Full-spectrum intelligence combines cumulative intelligence with all of the capabilities so far discussed in this section. Although it is already an extensive list, it may very well still be incomplete in some ways, and possibly incorrect in other ways.

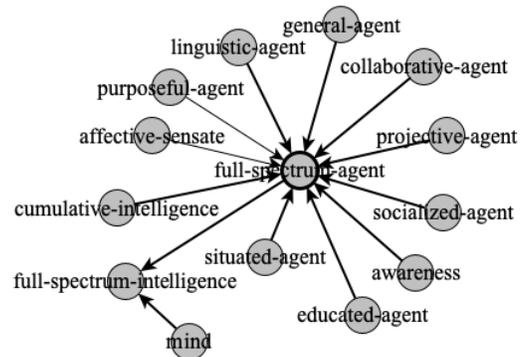

76. A *full-spectrum agent* has cumulative intelligence (goals, memories, rationality, and life-long learning) and is situated, aware, projective, general, affective, purposeful, linguistic, collaborative, socialized, and educated.

Figure 17: Dependency structure for Full-spectrum Intelligence.

- Looks to social and biological – particularly neural – sciences, as well as artificial (general) intelligence, for inspiration.
- May be realized at various levels of approximation.

77. *Full-spectrum intelligence* is what is yielded by the mind of a full-spectrum agent.

**1.4 Intelligence**

Here, the notion is that there is not a single idea that existing definitions of intelligence are striving to capture. Instead, a good part of the cacophony among such definitions arises from whether they implicitly concern immediate, cumulative, or full-spectrum intelligence. The alternative is to consider intelligence as a whole disjunctively, as concerning any one of these three. Beyond this, there is also a range of approximations around these three levels that contributes to inducing a broad space of possibilities for intelligence.

The analysis here starts with a definition of intelligence and several related concepts (Section 1.4.1) before following up with its relationship to a variety of existing defintions of intelligence (Section 1.4.2) and to several theories concerning multiple intelligences (Section 1.4.3).

*1.4.1 Defining Intelligence and the Intelligence Space (Figure 18)*

78. *Intelligence* can be immediate, cumulative, or full spectrum.

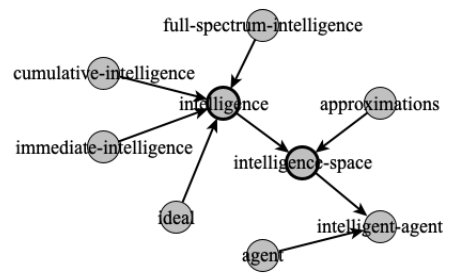

Figure 18: Dependency structure for Intelligence and the Intelligence Space.

79. The *intelligence space* comprises the three levels in the definition of intelligence plus a space of approximations around these levels.
    - This leaves open the possibility of a wide range of different forms of intelligence. However, since the definition of full-spectrum intelligence is highly influenced by our current understanding of human intelligence, as well as what is now known about artificial intelligence, this space may ultimately be viewed as skewed with respect to the space of all possible forms of intelligence.

80. An *intelligent agent* is an agent with a mind that sits somewhere in the intelligence space.
    - However, the approximation must not be so extreme that what is left is too minimal.

*1.4.2 Relationship to Existing Definitions of Intelligence*

It is worth taking a moment to consider the relationship of the definition of intelligence developed here to existing definitions, focused particularly on how they align with the hierarchy of levels laid out here. We can start with the first paragraph in Wikipedia [35] on intelligence:

"Intelligence has been defined in many ways: the capacity for abstraction, logic, understanding, self-awareness, learning, emotional knowledge, reasoning, planning, creativity, critical thinking, and problem-solving. More generally, it can be described as the ability to perceive or infer information, and to retain it as knowledge to be applied towards adaptive behaviors within an environment or context."

The first sentence approximates full-spectrum intelligence whereas the second maps onto either immediate or cumulative intelligence, depending on whether "adaptation" is intended to imply the flexibility to achieve goals or the addition of memories via learning.

Other examples of definitions aimed directly at immediate intelligence include:
- Intelligence comprises the ability to act rationally; that is, "does the 'right thing,' given what it knows." [12]

- "Intelligence measures an agent's ability to achieve goals in a wide range of environments." [36]

The second of these definitions does go one step beyond immediate intelligence to explicitly include the breadth of goals that yield a general agent. Given that this is a definition out of the world of AGI, this should not be a surprising addition.

Two example definitions aimed directly at cumulative intelligence, as quoted from Legg's webpage, are:
- "The capacity to acquire and apply knowledge." [37]
- "The capacity to learn or to profit by experience." (W. F. Dearborn, as quoted in [38]).

With respect to full-spectrum intelligence, the Turing test [39] explicitly tests whether an agent's behavior is indistinguishable from that of a human, and thus implicitly tests whether it shares many, if not all, of the discernable aspects of human-level, full-spectrum intelligence. The Newell test [24], a set of functional criteria for evaluating a theory of cognition, can also be viewed as implying a variant of human-level, full-spectrum intelligence. It is analyzed in more detail in Section 2.1, on human intelligence.

This mapping of a sampling of existing definitions of intelligence onto the three levels of the hierarchy provides prima facie evidence for their relevance, as does the work already mentioned as presaging the first and third levels, and the centrality of machine learning – the core of the second level – to the field of AI at this point. However, this need not imply that there may not still be other points within the overall space defined here that could usefully be called out explicitly, whether or not they track the nested hierarchy introduced here or are even part of a larger intelligence space not yet identified.

*1.4.3 Relationship to Theories of Multiple Intelligences*

As with unitary definitions of intelligence, it can be enlightening to consider theories of multiple intelligences through the lens of the definitions here. Such theories, although they tend to be controversial, are typically proposed as a counter to the assumption behind IQ tests that there is a single general intelligence factor – denoted $g$ – that can be measured by such tests. Here, two such theories are considered.

Sternberg's [40] *triarchic theory* of human intelligence includes *practical*, *creative*, and *analytical* intelligence. According to the definitions laid out here, practical intelligence emphasizes the contextual behavior found in situated agents; creative intelligence emphasizes the ability to come up with new ideas that is found in projective agents; and analytical intelligence emphasizes the ability to reason, understand, and achieve goals that is found in immediate intelligence plus aspects of awareness. Thus we see immediate intelligence, plus three particular aspects of full-spectrum intelligence: situated, projective, and aware.

Gardner's [41] theory of multiple intelligences is broader. It includes *musical*, *visual-spatial*, *linguistic*, *logical-mathematical*, *bodily-kinesthetic*, *interpersonal*, and *intrapersonal*, although even more were later added to this list. A number of these types of intelligence might best be viewed simply as situatedness with respect to particular kinds of environments and memories. However, three map more directly onto the definitions for aspects of full-spectrum-intelligence: linguistic intelligence maps onto linguistic beings; interpersonal intelligence maps onto collaborative, socialized agents; and intrapersonal intelligence maps onto self-awareness.

The net result of these two very brief analyses is that theories of multiple intelligences may not need to be considered as something very different from what is provided by the intelligence space here. Instead they seem to nest naturally within it.

# 2 Natural and Artificial Intelligence

This section starts with a particular form of natural intelligence – human intelligence – as an approximation to full-spectrum intelligence (Section 2.1) before widening the focus a bit to consider humanlike intelligence (Section 2.2). Artificial intelligence is then considered as both a form of intelligence that is distinct from natural intelligence and as a field of study (Section 2.3), before wrapping up with consideration of cognitive science as an interdisciplinary field that spans aspects of both natural and artificial intelligence (Section 2.4). All of these forms of intelligence can be considered as spanning regions of the intelligence space delineated in Section 1.4.1.

## 2.1 Human Intelligence (Figure 19)

Human intelligence is defined here as a form of natural intelligence that arises from human minds, as implemented by human brains (Section 2.1.1). This then enables defining the notion of a human cognitive architecture and the mapping of the Newell test for such architectures onto full-spectrum intelligence (Section 2.1.2).

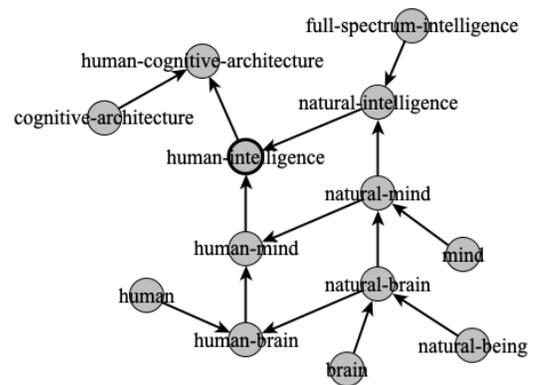

Figure 19: Dependency structure for Human Intelligence.

### 2.1.1 Defining Human Intelligence

81. A *natural brain* is the brain of a natural being.
82. A *natural mind* is a mind implemented by a natural brain.
83. *Natural intelligence* approximates full-spectrum intelligence via a natural mind.

84. A *human brain* is the natural brain of a human.
85. A *human mind* is a natural mind implemented by a human brain.
86. *Human intelligence* is a form of natural intelligence yielded by a human mind.
87. A *human cognitive architecture* is the architecture of a human mind.

### 2.1.2 The Newell Test (Figure 20)

The Newell test [24] comprises twelve criteria for a human cognitive architecture distilled down from two earlier lists [22, 23]: (1) flexible behavior, (2) real-time performance, (3) adaptive behavior, (4) vast knowledge base, (5) dynamic behavior, (6) knowledge integration, (7) natural language, (8) consciousness, (9) learning, (10) development, (11) evolution, and (12) (human) brain. The question of interest here is how these criteria relate to the definition of intelligence introduced in Section 1, and thus how to understand the Newell test in terms of this definition. Given that the Newell test was part of the input in developing the notion of full-spectrum intelligence, a close relationship should be expected, but a deeper analysis of this relationship can still yield useful insight into both.

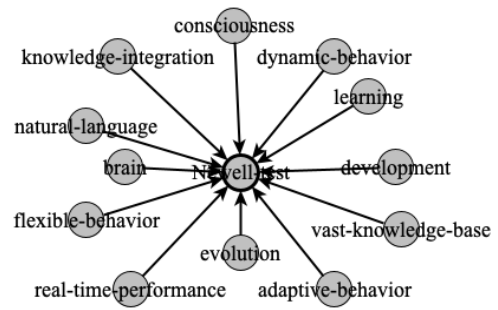

Figure 20: Dependency structure for the Newell test.

To start, we'll immediately eliminate criteria 10-12 from consideration. All three focus on natural agents, with the last being even more specific to humans. As criteria for a human cognitive architecture, these of course remain relevant, and earlier definitions in the overall cascade here do consider them, but the ultimate definition of intelligence developed in Section 1 is deliberately neutral as to whether it concerns natural versus artificial minds.

The remaining nine criteria can all be mapped onto aspects of full-spectrum intelligence. Criteria 3, 4, and 6 map onto immediate intelligence while criteria 4 and 9 map onto cumulative intelligence. The remainder map onto additional aspects of full-spectrum intelligence: criteria 2 and 5 map onto situated agents; criterion 8 maps onto awareness; criterion 3 maps onto projective agents; criterion 1 maps onto general agents; criterion 7 maps onto linguistic beings; and criterion 4 maps onto educated agents. Criterion 4, of a vast knowledge base, thus turns out to span all three levels of intelligence while criterion 3, of adaptive behavior, spans two of them.

Full-spectrum intelligence thus does subsume the type-neutral aspects of the Newell test but, in the reverse direction, two features of full-spectrum intelligence turn out to be missing from the Newell test: the notion that something "truly matters," as is central to both affective sensates and purposeful agents; and the interpersonal/social aspects that are central to collaborative and socialized agents. Full-spectrum intelligence, at least as defined here, proposes a more comprehensive list for what is necessary for intelligence in general.

## 2.2 Humanlike Intelligence (Figure 21)

The notion of humanlike intelligence assumes that there exists a compact region of the intelligence space around human intelligence that encompasses forms of both natural and artificial intelligence. The latter, while distinct from human intelligence, must be similar enough to it that a single model, at a suitable level of abstraction, is sufficient to characterize the whole region. The concept, although implicitly in existence for some time, became explicit in the development of the Common Model of Cognition – née the Standard Model of the Mind [42] – a community consensus concerning what must be in an architecture for humanlike minds.

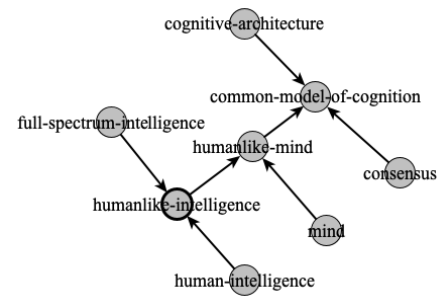

Figure 21: Dependency structure for Humanlike Intelligence.

88. *Humanlike intelligence* comprises forms of full-spectrum intelligence that are akin to human intelligence.
    - This definition is neutral with respect to whether natural or artificial intelligence is implicated.
89. A *humanlike mind* is a mind that exhibits humanlike intelligence.

90. A *common model of cognition* is a consensus concerning what must be in a cognitive architecture for a humanlike mind.
    - It explicitly accepts falling short of completeness where no consensus exists.

## 2.3 Artificial Intelligence (AI) (Figure 22)

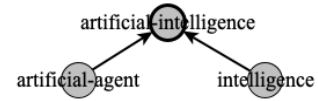

The range of existing definitions of *artificial intelligence (AI)* is at least as confusing as that for intelligence itself, with existing definitions tending to either focus on the perceived gap between human intelligence and what computers can currently do (e.g., [43, 44]) or some explicit notion of the aspects required for full-spectrum (artificial) intelligence (e.g., [45, 46]). However, with a definition of intelligence in hand, a very straightforward definition of AI becomes possible.

Figure 22: Dependency structure for Artificial Intelligence.

91. *Artificial intelligence* is intelligence in artificial agents.

This simple definition, while cutting through much of the murk, unfortunately still fails to capture much of the diversity found across the field (Figure 23). Although part of this diversity is due to what level of intelligence – immediate, cumulative, or full-spectrum – is being considered, there are additional sources that also need to be acknowledged, such as:
- *Humanness* varies over arbitrary forms of intelligence versus humanlike intelligence.
- *Generality* varies over narrow versus general agents.
  - A *narrow agent* only considers a few goals.
- *Completeness* varies over whether the focus is on individual aspects of full-spectrum intelligence versus all of it.
  - For the former, more variation results from which aspect(s) are the focus.
- *Embodiment* varies over agents, bots, virtual humans, robots, androids, or cyborgs.
- *Technology* varies over neural, symbolic, or hybrid.
  - Along with many possible variations within each of these broad categories.

92. The *original grand vision* for the field, coming out of the 1956 Dartmouth workshop, could be said to comprise complete, full-spectrum intelligence, with a bias towards symbolic technologies.
    - The desire for general agents is implied by the overall focus on complete full-spectrum intelligence.
    - Even at the beginning there were differences concerning forms of embodiment and degrees of humanness sought.
    - Much of AI's history has, however, focused on other points in this diverse space of possibilities rather than on this original grand vision.

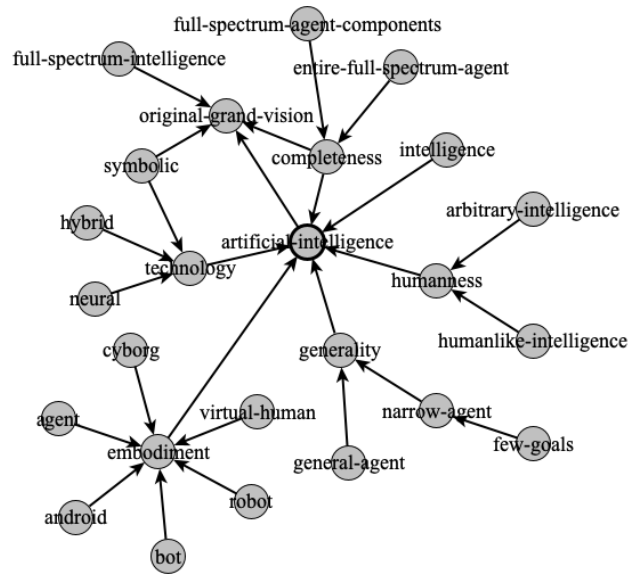

Figure 23: Dependency structure for diversity within the field of AI, including the original grand vision.

## 2.4 Cognitive Science (Figure 24)

*Cognitive science* is an interdisciplinary field that according to a broad view studies humanlike intelligence, although its zeitgeist centers on human intelligence, with artificial intelligence only then relevant to the extent that it assists in understanding human intelligence.

93. The field of *cognitive science* studies human (or humanlike) minds.

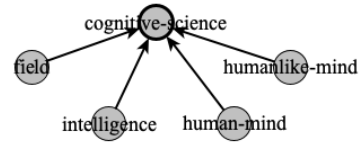

Figure 24: Dependency structure for the field of Cognitive Science.

According to either perspective on cognitive science, the field is grounded in multiple disciplines (Figure 25), each according to how it can contribute:
- Artificial intelligence: Artificial minds
- Philosophy: Questions, concepts, and formalisms
- Psychology: Data and theories about human minds
- Linguistics: Study of language structure and use
- Neuroscience: Data/theory that ground a human mind in a human brain
- Anthropology: Evolutionary and societal/cultural aspects of cognition.
- Sociology: Data/theory on natural societies
- Computer science: Study and construction of computational systems.

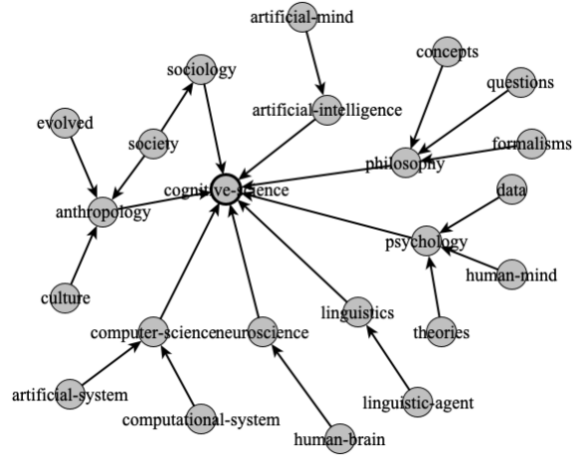

Figure 25: Dependency structure for the interdisciplinarity of Cognitive Science.

# 3 Exploring Implications

This section makes a beginning at exploring four more advanced topics for which relevant groundwork has been laid via the definitions in the preceding two sections: the singularity, as it applies to intelligence; generative AI, with a particular focus on large language models (LLMs); ethics, with a particular focus on AI safety; and intellectual property, with a particular focus on the possibility of AI generation of it. Each is, however, only a start, and each may introduce its own sources of controversy.

## 3.1 Intelligence Singularity (Figure 26)

The singularity, when considered in the context of intelligence, is to occur when AI agents are able to design new AI agents that are "smarter" than themselves, leading to an exponential growth in intelligence that far surpasses human intelligence [47]. This concept is considered in this section.

94. *Human-level intelligence* provides an approximation to intelligence that is comparable to what is provided by human intelligence.
    - This can be with respect to any of the three levels of intelligence.
    - This differs from humanlike intelligence in being about the degree of approximation to, rather than the form of, intelligence.
95. *Superhuman intelligence* provides a better approximation to some level of intelligence than is found in human intelligence.

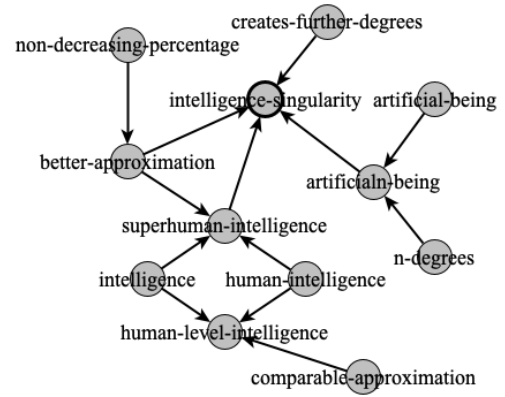

Figure 26: Dependency structure for the Intelligence Singularity.

96. An *artificial$^n$ being* is a being that is artificial to some – generically *n* – degrees.
    - An artificial$^0$ being is natural.
    - An artificial$^1$ being is artificial; that is, has been created by a natural being.
    - An artificial$^2$ being has been created by an artificial being.
        - This may be a new being with a different architecture or merely a modification of its own architecture.
    - An artificial$^n$ being has been created by an artificial$^{n-1}$ being.
    - For simplicity, the term *artificial* may be used for any $n>0$ when there is no need to indicate the specific non-zero degree or when the degree of artificiality is either unclear or mixed.

97. The *intelligence singularity* occurs when all of the following hold:
    - An artificial$^n$ *being* with $n>0$ attains superhuman intelligence.
    - It creates further degrees of artificialness in which each degree provides a better approximation to intelligence.
    - Better is measured in terms of a percentage that is non-decreasing.
        - This yields the hypothesized exponential increase in intelligence.

Stated this way, it should be clear what this form of singularity is, but it is not at all clear – at least to me – that there is sufficient headroom between the approximation provided by human-level intelligence to any of the levels of intelligence identified in this article that exponential improvement beyond human-level intelligence is possible, or at least in a manner that goes beyond a small number of degrees.

**3.2 Generative AI**

Generative AI produces new content – whether text, images, video, etc. – based on learning of parameters in a neural network from existing content. The most impactful form of it at this point comes in the form of *large language models (LLMs)* that learn to predict the next word in a sequence of words from previous words in it (e.g., [29]). It has come as rather a revelation that a technology initially developed simply as a means of rating alternative sentences generated by speech-to-text (i.e., speech recognition) and text-to-text (i.e., translation) systems in terms of how well they match the regularities/structure of the target language, become generators of language that, when training is sufficiently scaled up, captures in a useful and useable form much of the content of the textual datasets from which they are learned.

This section is focused in particular on LLMs. Their recent successes are due to algorithms that enable prediction from sequences of previous words of unbounded length (e.g., [48]) plus the use of vast textual datasets and the massive amounts of computing required to process them (e.g., [29]). The specific question asked here is what can be understood about them based on the definitions developed in Sections 1 and 2. In answering this question, only the basic capabilities of an LLM are assumed, without any additional architectural enhancements – whether reinforcement learning [49] or other mechanisms (e.g., [50]) – that may significantly alter the analysis.

Understanding LLMs in terms of the previous definitions is best done over multiple steps. The first two involve a basic definition (Section 3.2.1) followed by an analysis of the components that make up an LLM at a bit more detailed level (Section 3.2.2). LLMs are then mapped onto the trilevel hierarchy of intelligence (Section 3.2.3) before being reinterpreted in terms of the alternative perspective provided by the spectrum from natural to artificial intelligence (Section 3.2.4).

### 3.2.1 *Defining an LLM* (Figure 27)

98. *Pretrained parameters* are the parameters of an artificial system that are (machine) learned via pretraining.

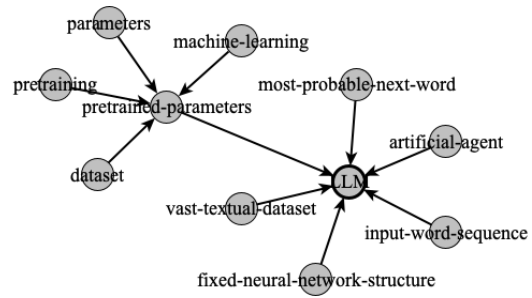

99. An *LLM* is an artificial agent that uses a fixed neural network structure plus parameters that are pretrained over a vast textual dataset to generate the most probable next word in a sequence given prior words in the sequence.

Figure 27: Dependency structure for an LLM.

### 3.2.2 *Analyzing the Components of an LLM* (Figure 28)

100. *LLM memories* comprise parameter values and the input word sequence.

101. An *LLM learning algorithm* is a machine learning algorithm that pretrains parameters in the network structure.

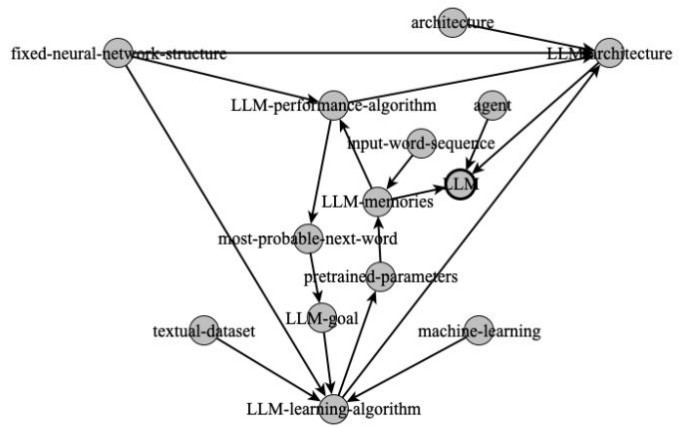

102. An *LLM performance algorithm* generates the most probable next word given its memories and the fixed neural-network structure.
103. An *LLM architecture* comprises a fixed neural-network structure plus an LLM learning algorithm and an LLM performance algorithm.

Figure 28: Dependency structure for LLM Components.

104. An *LLM goal* is to produce the most probable next word in the output sequence.

- This goal is explicit in the learning algorithm but implicit in the performance algorithm.
- An LLM may have many implicit goals, based on requests expressed in input sequences. But these implicit goals are just words in the input that lead to behavior that can be interpreted as teleological without being set as goals to be achieved.

*3.2.3 Understanding LLMs in Terms of the Intelligence Hierarchy*

An LLM exhibits immediate intelligence, but only with respect to its explicit goal.
- Rationality with respect to implicit goals is more a matter of chance, being a side effect of how memories are encoded in service of the explicit goal.

An LLM approximates cumulative intelligence.
- It learns extensively but utilizes pretraining rather than lifelong learning and learns only from what it is spoon-fed rather than from its interactive experiences.

An LLM approximates full-spectrum intelligence.
- It approximates a situated agent.
  - It is a bot in an environment of word sequences, and to a large extent behaves appropriately within the constraints of those sequences but it does not do so within the constraints of the larger reality and social environment within which it exists.
- It very minimally approximates awareness.
  - It has models without meaning, understanding, or awareness.
    - Despite that its memories might yield pseudo-awareness in terms of outputs that give the impression that it has these other attributes of awareness.
- It approximates a projective agent.
  - The architecture does not directly support this, but the memories do enable limited forms of it that do not appear to reach human level.
  - What it also doesn't have is the pervasive ability to project the consequences of its actions before executing them, and thus can't self-filter actions based on their projected consequences.
- It provides a very good approximation to a general agent.
  - With respect to individual humans, it can exhibit superhuman generality in some ways while simultaneously falling short of what most people can do in others. It is thus unhumanlike in its generality.
  - It falls short of what humanity as a whole can do due to its other limitations.
- It is not an affective sensate.
  - It has no feelings as nothing truly matters to it.
- It is not a purposeful agent.
  - It has no purposes as nothing truly matters to it.
- It approximates a linguistic being.
  - It is a communicative being that does not understand the words it uses and thus has no language and remains far below human level as a linguistic being.
- It is a minimal approximation to a collaborative agent.
  - It has no shared goals other than what is implicit in the word sequences upon which it operates.
- It is a minimal approximation to a socialized agent.

- It has no awareness but may be trained on materials that reflect aspects of its society's culture.
- It is a minimal approximation to an educated agent.
  - The extent to which it is educated is a function of how curated its training material is and how faithfully its learning algorithm reflects this material in its memories.

Given this analysis, it is clear that LLMs miss a number of key aspects of full-spectrum intelligence, and thus also key aspects of the human psyche that make us what we are and make us, at least most often, reliable partners in the human community (a topic that is revisited more generally in Section 3.3). From this analysis, there is also no reason to believe that LLMs are a major step toward the intelligence singularity. The only way in which they are superhuman is in generality, and even there it is of a restricted sort.

I have at times considered LLMs metaphorically as akin to the monster from the movie *Forbidden Planet* – an amplified projection of Dr. Morbius's *id*, unconstrained by his *ego* or *superego*. An LLM instead has at its core what is effectively an amplified but unconstrained *subconscious* (or, alternatively, body of memories). It is amplified in the sense of being broader in what it knows than the subconscious of a typical human, and in some ways also in the range of capabilities it can exhibit. It truly does seem superhuman in this breadth. Yet it is not superhuman in its depth – for example, in either rationality or projection – and thus does not seem a promising candidate for an artificial agent that can itself create better artificial agents.

*3.2.4 Understanding LLMs in Terms of the Spectrum from Natural to Artificial Intelligence*

An LLM is an instance of artificial intelligence.
An LLM can exhibit humanlike intelligence with respect to its behavior.
- There are reports that LaMDA, a particular LLM, has passed the Turing test [51]; as well as other intriguing reports (e.g., [52]).
- This is true even given the ways LLMs fail to yield full-spectrum intelligence.
  - Some of the deficiencies may just not show up in conversation.
  - Other deficiencies may be partially ameliorated by the use of memories that enable acting as if the LLMs embody various aspects that they don't truly possess.

An LLM may exhibit humanlike intelligence with respect to its architecture.
- Beyond the abstract mapping between artificial neural networks and networks of neurons in human brains, studies are showing a link between the key technology underlying most LLMs – transformers [48] – to data from the human brain (e.g., [53, 54]).

## 3.3 Ethics

The overall goal of this subsection is to begin a process of understanding how it might be possible to assign rights and responsibilities to objects based on their properties. It arose from a sense over the years that there is a difference between AIs as tools versus AIs as organisms that deeply impacts how safety can and should be approached for them, as well as what rights and responsibilities they should have. However, the notion of what an organism is when it came to artificial systems was always rather vague and how to principledly assign rights and responsibilities based on properties of tools versus organisms has remained opaque. Although what was previously referred to as an organism corresponds to some form of agent according to the definitions in Section 1, the body of definitions provided so far does begin to provide some leverage in considering safety, rights, and responsibilities. This is where this section leads.

If done right, such an enterprise should ultimately deal appropriately with both natural and artificial beings, both humans – including various forms of human neurodiversity – and other types of animals, and both terrestrial and (hypothetical) extraterrestrial beings. With respect to artificial beings, this topic clearly relates to AI ethics. However, the dominant thread there, concerned with ensuring the safe impact of AI systems on humanity, is just one piece of the overall picture, and as conventionally construed seems more directly applicable to objects that can be viewed as tools rather than to those with higher levels of intelligence.

Out of all of the topics covered in this article, those around ethics seem the most likely to be controversial. Yet, an even more essential warning is also necessary at this point: due to the potential implications of any proposed ethical system – and the disastrous results that may come from not fully understanding the implications early enough – great care should be taken with any potential applications of these ideas. Of most concern would be anything that might appear to suggest reducing the rights currently ascribed to any subsets of humanity. Still, the goal is to at least make a start at understanding how properties of objects might factor into ethical decisions concerning them in a more nuanced fashion than the gross dichotomies mentioned in the previous paragraph.

This section first introduces the notion of a tool and concepts around work, before moving on to alignment, intent, property, and slavery. Based on these notions it then explores the topic of AI safety before wrapping up with a discussion of rights and responsibilities that builds upon the basic definitions of these terms in Section 1.3.9, but which attempts baby steps toward understanding how they might relate to properties of objects.

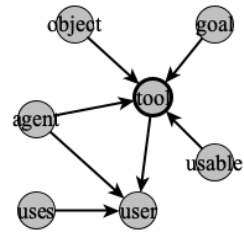

Figure 29: Dependency structure for a Tool.

### 3.3.1 Tool (Figure 29)
105. A *tool* is an object useable by an agent in service of its goals.
106. A *user* is an agent that uses a tool.

### 3.3.2 Work (Figure 30)
107. A *contractor* is an agent that achieves a goal for pay.
108. A *principal* is an agent that pays a contractor to achieve a goal.

109. A *subordinate* is an agent that is under an authority in some context.
    - The authority has some ability to determine the rights and responsibilities of the subordinate.

110. An *employee* is a subordinate contractor.
111. An *employer* is a principal who is an authority over an employee.

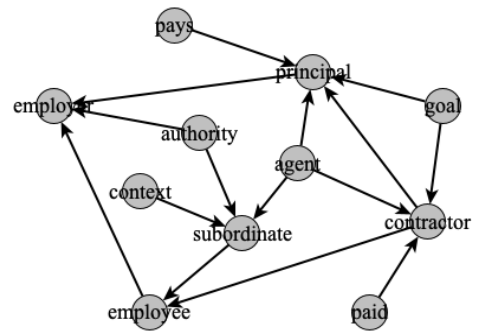

Figure 30: Dependency structure for Work.

### 3.3.3 Alignment (Figure 31)
112. *Incompatible goals* cannot be achieved simultaneously.
113. *Alignment* is the process of resolving incompatible goals across agents.

- This idea stems from *agency theory*, which actually concerns the other notion of agents, as representatives (see, e.g., [55]).
114. *AI alignment* is the process of alignment between AI and human goals (see, e.g., [56]).
    - It occurs by adjusting AI goals and the flexibility AIs have in achieving them.

115. *Modification* alters an existing agent so that it is subordinated and/or aligned.
    - *Training* provides experience to an agent that enables modification via learning.
    - *Programming* directly modifies an agent's structures and processes.
116. *Breeding* yields new agents that are (more) subordinated and/or aligned from one or more existing agents that may not be as fully subordinated and/or aligned.
117. *Governance* adds a governor (Section 3.3.8), created in some fashion, to oversee the agent.

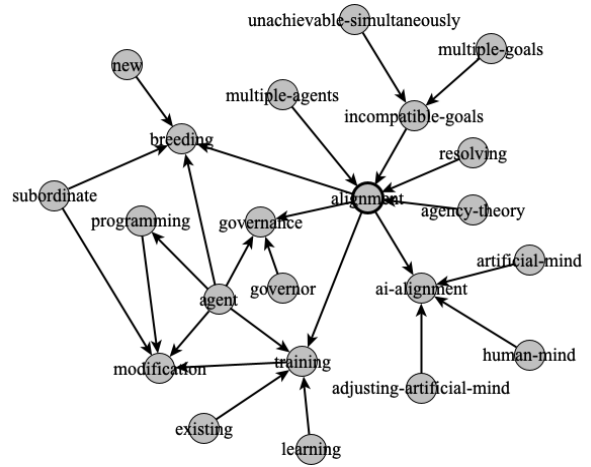

Figure 31: Dependency structure for Alignment.

### 3.3.4 Intent (Figure 32)
118. An *intent* is a purpose of a self-aware agent.
119. An *intentional agent* has intents.[4]

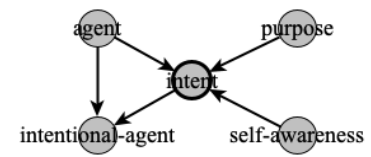

Figure 32: Dependency Structure for Intent.

### 3.3.5 Property (Figure 33)
120. An agent is the *owner* of an object if it has authority over it and has the right to sell, damage, or destroy it.
121. *Property* is an owned object.

### 3.3.6 Slavery (Figure 34)
This is a particularly awkward topic for an article on intelligence, and this brief discussion won't do it justice, but a key concern with overfocusing on AI alignment is the possibility of turning AIs into slaves. This concern is discussed further in Section 3.3.7.
122. A *slave* is an intentional subordinate that is property.
123. A *slave owner* owns and is an authority over slaves.
    - Effectively becomes the user of slaves as tools.

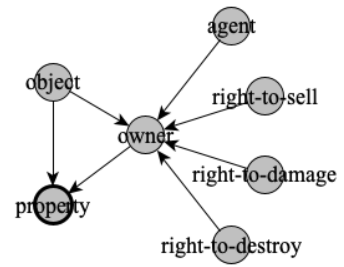

Figure 33: Dependency structure for Property.

### 3.3.7 AI Safety (Figure 35)
124. A *hazard* is a potential danger to an object.
125. *Safety* concerns the protection of an object from a hazard.
126. A *safe tool* is usable without hazard to its user.

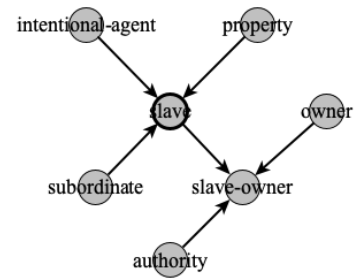

Figure 34: Dependency structure for Slavery.

---

[4] This notion differs from Dennett's [57] conceptualization of intentional systems, which is closer to what are termed agents here.

127. An *AI hazard* is a hazard from an artificial intelligence.
128. *AI safety* concerns removing AI hazards for humans.
    - AI alignment does this by making them into safe tools for humans.
        - But it is important not to turn them into slaves in the process.
    - An alternative is extending AI into *responsible members of society*.
        - Involves including more aspects of full-spectrum intelligence.
            - For LLMs (Section 3.2), this most particularly involves including awareness, collaboration, socialization, and education.
                - Purposefulness may be needed as well.

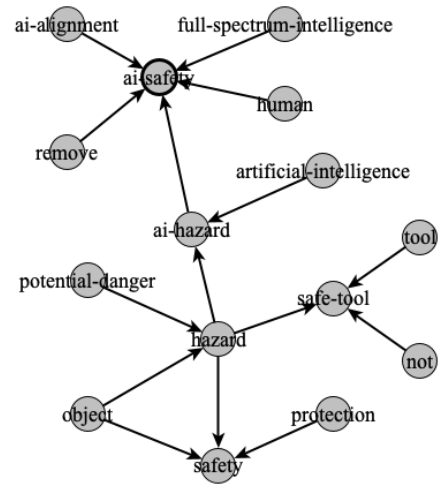

Figure 35: Dependency structure for AI Safety.

This bipartite perspective on AI safety captures the essence of the tools-versus-organisms dichotomy that originally motivated this section. In general, I have two concerns about the alignment approach to AI safety. One has just been mentioned, of when it would involve turning AIs into slaves, and possibly then lead to the kinds of dystopias imagined in science fiction involving robot revolutions. The other is whether agents that are sufficiently intelligent and that operate in sufficiently open worlds can ever be made safe through some form of pre-alignment.

A better alternative would seem to be to enable AI agents to do what people do; that is, to understand what they are considering doing in the context of what they know, what matters to them, and what they know matters to others and to society as a whole. Both of the above arguments trend towards making AI agents responsible members of society rather than trying to modify them into fully aligned tools. The accompanying question of appropriate rights and responsibilities for such agents is part of the broader discussion of what rights and responsibilities might belong to different types of objects, as begun in the next section.

### 3.3.8 *Rights and Responsibilities*

The basic definitions of rights and responsibilities can be found in the earlier discussion of socialized agents (Section 1.3.9). The material here goes beyond simply the social aspects of agents in an attempt to begin a broader discussion that encompasses arbitrary objects. Following some preliminary discussion of rights and responsibilities new definitions concerning them are introduced. To simplify things, the dependency structure here is divided into one for rights (Figure 36) and one for responsibilities (Figure 37), albeit with parallel structure across the two.

The ultimate question here is what kinds of objects should have what kinds of rights and responsibilities, and on what aspects of those objects should they depend. A number of prior efforts have taken up the general question of whether robots should have rights (e.g., [58]), including ones making explicit arguments either for (e.g., [59]) or against (e.g., [60]); or arguing that they should have responsibilities rather than rights [61]. At least one such effort [62] attempted to dig deeper, in discussing whether membership in a moral community should depend on rationality (i.e., immediate intelligence), sentience (i.e., awareness), or interests (i.e., goals/purposes).

Rights may be relevant to any object.
- Consider, for example, how a culturally significant painting or building might be ascribed certain rights.

Responsibilities may be relevant to any agent.
- Without goals, there is no effective way to take on responsibilities.

As discussed in Section 1.3.9, both rights and responsibilities derive from an authority.
- Such an authority may be an employer, an owner, a society, a god, a ruler, itself, or in principle any kind of agent.
- A document or abstract ethical system is not an authority in this sense, not being an agent, but can be the record of the rights and responsibilities earlier authorized by an agent.
- Nothing is intended to be implied here about how an agent becomes an authority or what it might mean for such an authority to be legitimate.

Both rights and responsibilities become more compelling when things "truly matter."
- This captures one of the essential implications of the notion of something truly mattering.
- It encompasses purposeful agents; and, for rights, it is also encompasses affective sensates.

Both rights and responsibilities become even more compelling for intentional agents.
- Such agents are aware of their rights and responsibilities and of the impact of these on what "truly matters" to them and others.

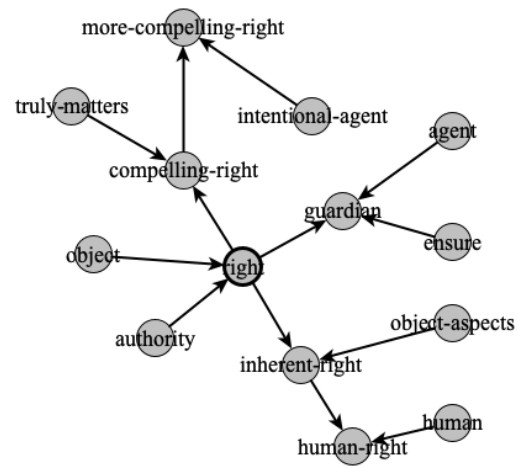

Figure 36: Dependency structure for Rights.

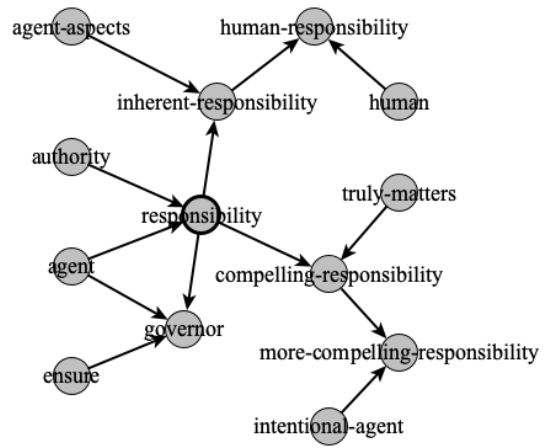

Figure 37: Dependency structure for Responsibilities.

129. An *inherent right* or *responsibility* derives from aspects of the object in question.
    - This assumes, even if only implicitly, that there is an authority that endows the object with this right or responsibility.
    - The notion of inherent rights can be considered as a descendant of the traditional notion of *natural rights*.
130. A *human right* or *responsibility* is an inherent right or responsibility that derives from being human.

131. A *guardian* is an agent that is responsible for the rights of an object.
    - An agent may be its own guardian.

- Guardians may also, for example, be relevant to the rights of plants, animals, ecosystems, children, some forms of neurally diverse adults, and objects of cultural significance.
- At some point guardians may also be relevant to some kinds of AI systems.

132. A *governor* is an agent responsible for ensuring that another agent upholds its responsibilities.[5]

Determining what inherent rights and responsibilities should arise as a function of different aspects of different types of objects is in my view critical for a general theory of ethics, but it is considerably beyond what can be attempted here. Still, examples of potentially relevant questions include:
- What rights should affective sensates have with respect to their feelings?
- What rights should purposeful agents have with respect to their purposes?
    - More minimally, what rights should agents have with respect to their goals?
- When, and how, is it okay to terminate or destroy objects of different types?
    - As a function of its feelings, purposes, goals, self-awareness, cultural value, or other aspects (including the role it plays in the feelings, purposes, or goals of others).
- What rights and responsibilities should objects have when it is unclear what their relevant aspects are?
    - This can be particularly relevant, for example, to questions concerning animal rights, where there is still much that we do not understand about them.
    - This may also be relevant, if not now then before too long, to consideration of complex, poorly understood AI systems.
- What rights and responsibilities should generative AIs in particular have as a function of their specific approximation of full-spectrum intelligence?
    - Should they be entitled to any rights given that nothing "truly matters" to them and that they do not appear to have any other aspects that strongly bear on this question?
    - Can they be trusted to take on some responsibilities given their status as agents?
        - Their lack of awareness, collaboration, socialization, and education all lead to critical questions concerning how well they may be able to take on many responsibilities.
- What rights and responsibilities should corporations have as a function of their particular approximation to full-spectrum intelligence?[6]
    - I won't attempt an analysis in any detail here, but does the fact that they may not have true purposes, and only one explicit goal – of maximizing profits – imply that they cannot be responsible members of society and thus should have their rights more constrained and their responsibilities appropriately governed?[7]

---

[5] In an early AI perspective on this, Arkin, Ulam & Duncan [63] propose a constraint-based agent as the governor for weaponized military robots.

[6] The notion that a corporation is in fact an AI has been suggested a number of times (e.g., [9, 64, 65]) although, according to Section 1.3.8, corporations – as human organizations – should actually be considered as hybrid agents.

[7] Through this lens, *environmental, social governance* (*ESG*) can be seen as one route toward making them more responsible members of society.

- Does the ability to be projective impact the level of responsibility that can be assumed?
    - Without projection, unanticipated situations can yield wildly inappropriate behavior.
- Should self-awareness on its own imply certain rights?

## 3.4 Intellectual Property

The question of whether an AI agent can be, or should be, acknowledged as an inventor, an author, an artist, or a discoverer raises significant issues for the assignment of intellectual property. This section attempts to resolve these questions by focusing on whether the agent is intentional, and thus self-aware (Section 1.3.2) and purposeful (Section 1.3.6).[8] This focus on intentionality could easily be controversial, as well as problematic given how unclear the notion of "truly matters" remains, but it does at least provide a potentially principled rationale for resolving such questions that is not simply based on whether or not the agent is human.

If this approach does prove to be too problematic or inappropriate in its implications, it will hopefully inspire follow-up attempts that manage better. As it is, though, it does imply that current AI agents should not themselves be considered creators of intellectual property. This conclusion builds up over definitions concerning inventors, artists, discoverers, and ultimately the idea of intellectual property itself.

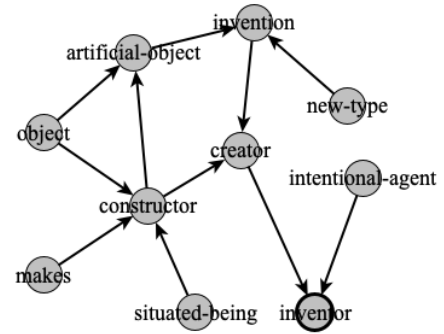

### 3.4.1 <u>Inventor</u> (Figure 38)
Here, a cascade of definitions build up from a constructor, to a creator, to an inventor.

133. A *constructor* is a situated agent that makes objects.
134. An *artificial object* is made by a constructor.
    - The concept of an artificial being was already introduced in Section 1.1.2. In terms of the definitions in this section, an artificial being is an artificial object that takes the form of a being and is constructed by a human.

Figure 38: Dependency structure for an Inventor.

135. An *invention* is a new type of artificial object.
136. A *creator* is a constructor of inventions.
137. An *inventor* is a creator that is an intentional agent.

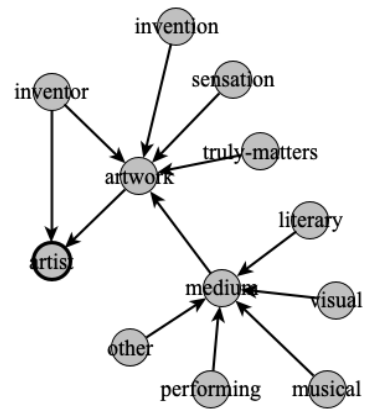

### 3.4.2 <u>Artist</u> (Figure 39)
There is no generally accepted definition of what art is, but Wikipedia [66] contains a useful characterization of "artworks (art as objects) that are compelled by a personal drive (art as activity) and convey a message, mood, or symbolism for the perceiver to interpret (art as experience). Art is something that stimulates an individual's thoughts, emotions, beliefs, or ideas through the senses."

Figure 39: Dependency structure for an Artist.

---

[8] The issue of generative AIs being trained over datasets that include intellectual property that belongs to others, while an important one in itself, is not considered here. The definitions laid out so far do not appear to provide leverage with respect to this particular issue.

This characterization provides a lot to work with, including implicitly considering that the art must "truly matter" to the artist in being "compelled by a personal drive."  It is less clear whether it assumes the art that results must "truly matter" to those sensing it but this will be assumed here.  Stimulation of emotions clearly involves this but whether stimulating beliefs or ideas does is iffier.  Yet, without this, generating any kind of information that truly matters to its producer could be considered art as long as it gets the recipient to simply think about it.

138. An *artwork* is an invention by an inventor that provides a sensation that "truly matters" to others.
  - The sensation may "truly matter" by relating to feelings or purposes.
  - An artwork may exist within any medium, including but not limited to *visual*, *performing*, *literary*, and *musical*.
139. An *artist* is an inventor of an artwork.

### 3.4.3 Discoverer (Figure 40)
140. A *discovery* is the finding of a previously unknown object.
  - There is a grey area in which it can be ambiguous as to whether something new has been found or constructed.  This is particularly subtle in areas such as mathematics, in a thread that goes back at least to Plato, who considered mathematical objects to be discovered rather than invented (see, e.g., [67]).
141. A *finder* makes a discovery.
142. A *discoverer* is a finder that is an intentional agent.

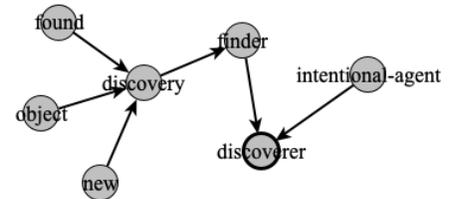

Figure 40: Dependency structure for a Discoverer.

### 3.4.4 Intellectual Property (Figure 41)
Given the prior definitions, it becomes straightforward to define intellectual property.

143. *Intellectual Property* (*IP*) is an invention or discovery that becomes property.

Artificial agents currently do not have purposes, and thus they can be neither inventors nor discoverers, although they may be creators or finders.  One implication of this is that they cannot by themselves engender intellectual property.  However, when used as a tool by an intentional agent, the artificial agent may be part of an act of invention or discovery by its intentional user.

In intermediate circumstances, where the artificial agent is more than a tool yet less than an intentional agent – and thus the act of creating or finding is sufficiently divorced from the intent of the agent's deployer – it becomes unclear whether there truly is an inventor or a discoverer, either individually or in conjunction.  The deployment of generative AI may fall within this unclear intermediate circumstance.

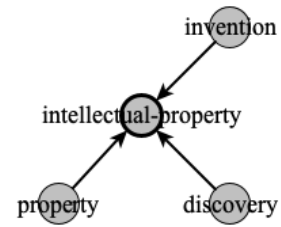

Figure 41: Dependency structure for Intellectual Property.

# 4 Conclusion

This article has amounted to a journey through 143 concepts/definitions that is illustrated by 41 graphs of dependencies among related concepts and associated commentary.  It has built up to a trilevel – immediate, cumulative, and full-spectrum – definition of intelligence that with associated approximations yields an intelligence space.  At a minimum, this space is intended to move forward our understanding of how to span both human and artificial intelligence, with mappings of these topics and several related ones onto this space.  This space may also yield a beginning at identifying a broader intelligence space that spans all possible forms of intelligence.

 The overall hope is that this analysis has brought some useful structure to how we think about intelligence and the different flavors of it, whether in terms of levels of intelligence and approximations to them or regions of the resulting space.  In turn, the hope is that this helps us see through the existing thicket of conflicting definitions of both intelligence in general and artificial intelligence more specifically, toward a common understanding of the relevant space of possibilities.  As mentioned at the beginning, this article is intended as only a first step in this direction rather than as a final word on the topic; so, if this does prove to be a useful direction, a variety of refinements and extensions would likely still be needed.

 The more advanced topics covered here are also first steps at best.  The analysis of the intelligence singularity raises questions about whether there is sufficient headroom above human-level intelligence for an exponential improvement over it.  The analysis of generative AI attempts to identify where large language models fit within the intelligence space.  There is uncertainty with respect to this due to how little we still understand about these models, but tentative conclusions can still be drawn in terms of their approximations to the three levels of intelligence.  The analysis of ethics has focused on questioning whether alignment is the right approach to AI safety versus developing AIs more into responsible members of society.  Concerns are included about both whether alignment is actually a workable solution and whether it risks ultimately turning AIs into slaves.  The analysis of intellectual property ultimately concludes that current AIs do not by themselves generate intellectual property, yet it does not rule out the possibility that future AIs, should they be intentional agents, might.

**Acknowledgement**

I would like to thank Volkan Ustun for helpful comments on a draft of this article.